\documentclass[runningheads]{llncs}
\usepackage[T1]{fontenc}
\usepackage{graphicx}
\usepackage{amsmath,amssymb}
\usepackage{booktabs}
\usepackage{multirow}
\usepackage{subcaption}
\usepackage{pifont}

\newcommand{\cmark}{\ding{51}}

\begin{document}

\title{URMF: Uncertainty-aware Robust Multimodal Fusion for Multimodal Sarcasm Detection}

\titlerunning{URMF for Multimodal Sarcasm Detection}

\author{
Zhenyu Wang\inst{1} \and
Weichen Cheng\inst{1} \and
Weijia Li\inst{1} \and
Junjie Mou\inst{1} \and
Zongyou Zhao\inst{1} \and
Guoying Zhang\inst{1}\thanks{Corresponding author}
}

\authorrunning{Z. Wang et al.}

\institute{
School of Artificial Intelligence, China University of Mining and Technology-Beijing, Beijing 100083, China\\
\email{zgy@cumtb.edu.cn}
}

\maketitle

\begin{abstract}
\label{sec:abstract}
Multimodal sarcasm detection (MSD) aims to identify sarcastic intent from semantic incongruity between text and image.
Although recent methods have improved MSD through cross-modal interaction and incongruity reasoning, most still treat modalities as equally reliable.
In real social media posts, however, text and images often differ in noise level and relevance, making deterministic fusion susceptible to noisy evidence and weakened incongruity cues.
To address this issue, we propose \textbf{U}ncertainty-aware \textbf{R}obust \textbf{M}ultimodal \textbf{F}usion (URMF), a unified framework for robust MSD.
URMF first injects visual evidence into textual representations through multi-head cross-attention, and then applies self-attention in the fused semantic space to enhance incongruity reasoning.
It models textual, visual, and interaction-aware representations as learnable Gaussian posteriors to estimate modality-specific uncertainty.
Based on the estimated uncertainty, URMF dynamically adjusts modality contributions during fusion to suppress unreliable evidence.
We further optimize the model with a unified objective that combines information bottleneck regularization, modality prior regularization, cross-modal distribution alignment, and uncertainty-driven contrastive learning.
Experiments on the public MSD and MMSD2 benchmarks show that URMF outperforms representative unimodal, multimodal, and MLLM-based baselines.
The results demonstrate that explicit uncertainty modeling can improve both accuracy and robustness in multimodal sarcasm detection.
\end{abstract}


\section{Introduction}
\label{sec:introduction}

Sarcasm is commonly expressed through a discrepancy between literal meaning and underlying intent.
With the prevalence of image-text posts on social media, multimodal sarcasm detection (MSD) has become an important task that requires modeling semantic incongruity across modalities.

Existing MSD methods have improved performance by enhancing cross-modal interaction and incongruity reasoning.
However, most of them still treat multimodal inputs as deterministic and equally reliable.
In real social media scenarios, text may be ambiguous or context-dependent, while images may be weakly related or even irrelevant.
Such unreliable modality information can be amplified by deterministic fusion and consequently weaken critical sarcasm cues.

Uncertainty estimation provides a natural way to address modality unreliability.
In particular, aleatoric uncertainty is suitable for multimodal learning because different modalities often exhibit different noise levels and reliability.
By estimating modality-specific uncertainty, the model can suppress unreliable evidence and perform more robust fusion.
Nevertheless, existing uncertainty-aware multimodal methods mainly focus on general robust fusion, paying limited attention to a key property of MSD: sarcasm cues are often formed after cross-modal interaction rather than contained in either modality alone.

To address this issue, we propose URMF, an \textbf{U}ncertainty-aware \textbf{R}obust \textbf{M}ultimodal \textbf{F}usion framework for MSD.
URMF first injects visual evidence into textual representations through cross-modal interaction and then performs self-attention in the fused semantic space to enhance incongruity reasoning.
It further models textual, visual, and interaction-aware representations with Gaussian uncertainty, and dynamically adjusts modality contributions according to their estimated reliability.
In addition, we introduce uncertainty-driven self-sampling contrastive learning and optimize the whole framework under a unified objective.

Our main contributions are summarized as follows:
\begin{itemize}
\item We propose URMF, a unified framework that combines cross-modal interaction with unimodal aleatoric uncertainty modeling for robust multimodal sarcasm detection.
\item We design an uncertainty-guided dynamic fusion mechanism and a unified training objective to improve robustness.
\item Experiments on the public MSD and MMSD2 benchmarks show that URMF consistently outperforms representative unimodal, multimodal, and MLLM baselines.
\end{itemize}

\section{Related Work}

\subsection{Multimodal Sarcasm Detection}

Multimodal sarcasm detection (MSD) aims to identify sarcastic intent by modeling semantic incongruity between text and image.
Early methods mainly relied on direct feature fusion~\cite{cai2019multi}, while later studies improved performance through graph-based reasoning, attention-based interaction, Transformer-based fusion, and explicit incongruity modeling~\cite{wei2024towards,yuan2025enhancing,wang2025rclmufn,tian2023dynamic,wei2025deepmsd,zhang2025incongruity,wang2024cross,guo2025multi,zhou2025semirnet,zhou2025ldgnet}.
However, most existing MSD methods still treat multimodal inputs as deterministic and equally reliable, with limited modeling of modality reliability.

\subsection{Uncertainty in Multimodal Learning}

Uncertainty estimation is widely used to improve robustness in deep learning.
In multimodal learning, aleatoric uncertainty is important because different modalities often exhibit different noise levels and reliability~\cite{kendall2017uncertainties}.
Recent studies show that uncertainty can serve as an adaptive signal for robust multimodal fusion, especially under corrupted or incomplete modalities~\cite{gao2024embracing}.
Unlike these general uncertainty-aware fusion methods, URMF incorporates uncertainty estimation into cross-modal incongruity modeling by constructing an interaction-aware latent modality and performing dynamic fusion based on post-interaction semantics.

\section{Method}
\label{sec:method}

\begin{figure*}[t]
    \centering
    \includegraphics[width=\textwidth]{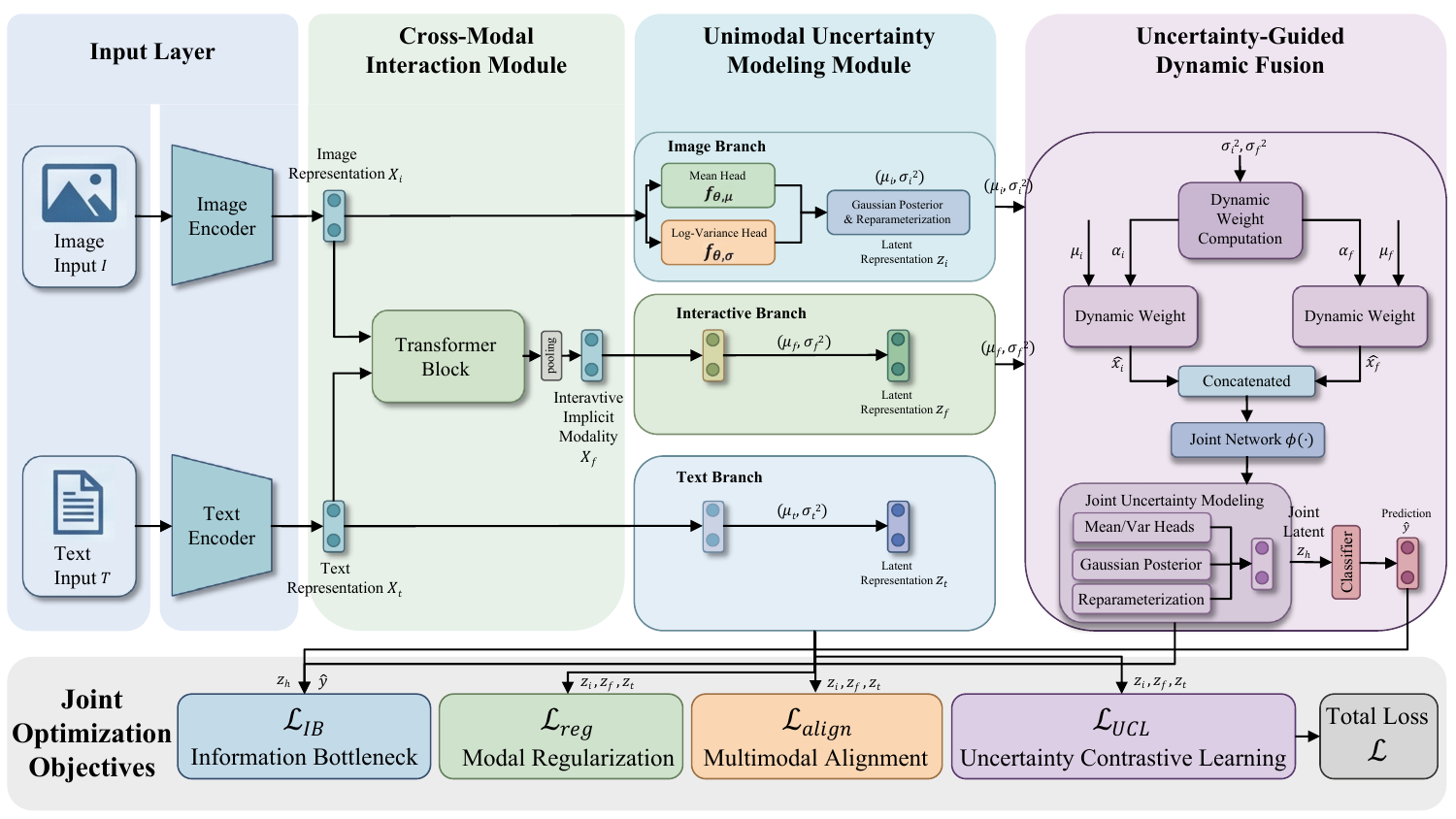}
    \caption{The overall framework of URMF.}
    \label{fig:framework}
\end{figure*}

\subsection{Overall Framework}

Given an image-text pair $(I,T)$, URMF first obtains an interaction-aware representation $X_f$, models textual, visual, and interaction-aware representations with aleatoric uncertainty, and performs uncertainty-guided fusion for prediction.
The framework is trained end-to-end.

\subsection{Cross-modal Interaction Module}
\label{3.2:cross_modal_interaction}

Unlike the common ``self-attention first, cross-attention later'' structure~\cite{tian2023dynamic}, we adopt a ``cross-modal interaction first, intra-modal reasoning later'' design.
Visual context is first injected into textual representations via cross-attention, followed by self-attention and a feed-forward network (FFN) in the fused semantic space.
Residual connections and layer normalization (LN) are used in each submodule.

Let $X_t \in \mathbb{R}^{n \times d_t}$ and $X_i \in \mathbb{R}^{m \times d_i}$ denote the textual and visual representations extracted by the text and image encoders.
The interaction module is defined as
\begin{align}
\label{eq:layer_overall}
X_t^{c} &= \mathrm{LN}\!\left(\mathrm{MHCA}(X_t, X_i) + X_t\right), \\
X_t^{s} &= \mathrm{LN}\!\left(\mathrm{MHSA}(X_t^{c}) + X_t^{c}\right), \\
X_t^{\prime} &= \mathrm{LN}\!\left(\mathrm{FFN}(X_t^{s}) + X_t^{s}\right),
\end{align}
where $\mathrm{MHCA}$ and $\mathrm{MHSA}$ denote multi-head cross-attention and multi-head self-attention, respectively.
The global interaction-aware representation is obtained by pooling:
\begin{equation}
\label{eq:pool_global}
X_f = \mathrm{Pool}(X_t^\prime).
\end{equation}

\subsection{Unimodal Aleatoric Uncertainty Modeling}
\label{3.3.un_modeling}

For the $k$-th sample, let $\mathbf{x}_m^k$ denote the representation of modality $m\in\{t,i,f\}$, corresponding to textual, visual, and interaction-aware modalities.
Following aleatoric uncertainty modeling~\cite{gao2024embracing}, each modality is represented as a Gaussian posterior, where the mean captures semantic information and the variance reflects modality uncertainty.

Two lightweight fully connected heads are used to predict the mean and log-variance:
\begin{equation}
\boldsymbol{\mu}_m^k = f^{m}_{\theta,\mu}(\mathbf{x}_m^k), \qquad
\log \boldsymbol{\sigma}_m^{k\,2} = f^{m}_{\theta,\sigma}(\mathbf{x}_m^k).
\end{equation}
The posterior distribution is defined as
\begin{equation}
p(\mathbf{z}_m^k \mid \mathbf{x}_m^k) = 
\mathcal{N}\!\left(\boldsymbol{\mu}_m^k,\ \boldsymbol{\sigma}_m^{k\,2}\mathbf{I} \right),
\end{equation}
and sampling is performed by reparameterization:
\begin{equation}
\mathbf{z}_m^k=\boldsymbol{\mu}_m^k+\boldsymbol{\sigma}_m^k\epsilon,\qquad 
\epsilon\sim\mathcal{N}(0,\mathbf{I}).
\end{equation}

\subsection{Uncertainty-guided Dynamic Fusion}
\label{3.4.fusion}

After uncertainty modeling, fusion weights are assigned according to modality uncertainty.
Since textual semantics have been injected into $X_f$, the final fusion is performed between the interaction-aware modality and the visual modality to reduce redundant textual information.

We first compute the mean variance as the scalar uncertainty:
\begin{equation}
\bar{\sigma}_m^{k\,2}
=
\frac{1}{D}\sum_{d=1}^{D}\boldsymbol{\sigma}_{m,d}^{k\,2},
\qquad m\in\{f,i\},
\end{equation}
where $D$ is the feature dimension.
Then, the dynamic fusion weight is computed as
\begin{equation}
a_m^k=
\exp\!\left(\frac{1}{\bar{\sigma}_m^{k\,2}+\varepsilon}\right),
\qquad
\alpha_m^k=
\frac{a_m^k}{\sum\limits_{s\in\{f,i\}}a_s^k},
\label{eq:alpha_uncertainty}
\end{equation}
where $\varepsilon$ is a numerical stability term.
The weighted modality contribution is obtained by
\begin{equation}
\hat{\mathbf{x}}_m^k=\alpha_m^k\,\boldsymbol{\mu}_m^k.
\label{eq:uncertainty_fusion}
\end{equation}

The contribution representations are concatenated and fed into a joint network.
The joint representation is further modeled as a Gaussian posterior and used for prediction:
\begin{align}
\mathbf{h}^k &= \phi([\hat{\mathbf{x}}_f^k,\hat{\mathbf{x}}_i^k]), \\
\boldsymbol{\mu}_h^k &= f_{\theta,\mu}^h(\mathbf{h}^k),
\qquad
\log\boldsymbol{\sigma}_h^{k\,2}=f_{\theta,\sigma}^h(\mathbf{h}^k), \\
\hat{y}^k &= \mathrm{softmax}(W\mathbf{z}_h^k+b).
\end{align}

\subsection{Training Objective}
\label{3.5:objective}

The total objective jointly optimizes task prediction, latent regularization, cross-modal alignment, and uncertainty-aware contrastive learning:
\begin{equation}
\mathcal L
=
\mathcal L_{\mathrm{IB}}
+
\lambda_{1}\mathcal L_{\mathrm{reg}}
+
\lambda_{2}\mathcal L_{\mathrm{align}}
+
\lambda_{3}\mathcal L_{\mathrm{UCL}}.
\label{eq:total_loss}
\end{equation}

The information bottleneck loss~\cite{alemi2016deep} consists of the task loss and KL regularization on the joint latent representation:
\begin{equation}
\mathcal L_{\mathrm{IB}}
=
\mathcal L_{\mathrm{task}}(y,\hat y)
+
\lambda_{\mathrm{IB}}\,
\mathrm{KL}\!\left(p(\mathbf{z}_h^k\mid \mathbf{h}^k)\,\|\,\mathcal N(0,\mathbf{I})\right),
\label{eq:lib}
\end{equation}
where $\mathcal L_{\mathrm{task}}$ is the cross-entropy loss.

The modality prior regularization stabilizes the textual, visual, and interaction-aware posteriors:
\begin{equation}
\mathcal L_{\mathrm{reg}}
=
\sum_{m\in\{t,i,f\}}
\mathrm{KL}\!\left(p(\mathbf{z}_m^k\mid \mathbf{x}_m^k)\,\|\,\mathcal N(0,\mathbf{I})\right).
\label{eq:lreg}
\end{equation}

To reduce cross-modal distribution shift while preserving modality-specific functions, we introduce an asymmetric alignment loss that uses the visual modality as the context anchor:
\begin{equation}
\mathcal L_{\mathrm{align}}
=
\mathrm{KL}\!\left(p(\mathbf{z}_t^k\mid \mathbf{x}_t^k)\,\|\,p(\mathbf{z}_i^k\mid \mathbf{x}_i^k)\right)
+
\mathrm{KL}\!\left(p(\mathbf{z}_f^k\mid \mathbf{x}_f^k)\,\|\,p(\mathbf{z}_i^k\mid \mathbf{x}_i^k)\right).
\label{eq:lalign}
\end{equation}

Finally, uncertainty-aware contrastive learning is constructed by sampling twice from each unimodal posterior.
For modality $m\in\{t,i,f\}$, the two samples from the same instance form a positive pair, while samples from other instances are treated as negatives:
\begin{equation}
\mathcal{L}_{\mathrm{UCL}}^m
=
-\log
\frac{
\exp\!\left(\mathrm{sim}\!\left(\tilde{\mathbf{z}}_{m}^{k},\,\mathbf{z}_{m}^{k}\right)/\tau\right)
}{
\sum_{k\ne k^{\prime}}
\left(
\exp\!\left(\mathrm{sim}\!\left(\tilde{\mathbf{z}}_{m}^{k},\,\mathbf{z}_{m}^{k}\right)/\tau\right)
+
\exp\!\left(\mathrm{sim}\!\left(\tilde{\mathbf{z}}_{m}^{k},\,\mathbf{z}_{m}^{k^{\prime}}\right)/\tau\right)
\right)
},
\end{equation}
\begin{equation}
\mathcal{L}_{\mathrm{UCL}} = \sum_{m \in \{t,i,f\}} \mathcal{L}_{\mathrm{UCL}}^m,
\end{equation}
where $\tau$ is the temperature coefficient.
Since the perturbation is controlled by $\boldsymbol{\sigma}_m^k$, this loss improves robustness by exploiting uncertainty-induced local variations.

\section{Experiments}

\subsection{Experiment Setting}

We evaluate URMF on two public multimodal sarcasm detection benchmarks, MSD~\cite{cai2019multi} and MMSD2~\cite{qin2023mmsd2}.

For both datasets, we follow the official train/validation/test splits and preprocessing procedures.
We report Accuracy, Precision, Recall, and F1-score as evaluation metrics.

For feature extraction, we use RoBERTa as the text encoder and ViT as the image encoder.
Unless otherwise specified, all experiments are conducted on a single NVIDIA GeForce RTX 4090 GPU.
The key hyperparameters are listed in Table~\ref{tab:hyperparams_symbol}.

\begin{table}[t]
\centering
\small
\setlength{\tabcolsep}{7pt}
\begin{tabular}{l l l}
\hline
\textbf{Symbol} & \textbf{Value} & \textbf{Description} \\
\hline
$n$ & $100$ & Maximum length of text tokens \\
$m$ & $49$ & Number of image patches \\
$d_t$ & $768$ & Text embedding dimension \\
$d_i$ & $768$ & Image embedding dimension \\
$\lambda_{\mathrm{IB}}$ & $10^{-3}$ & Weight of the KL term in $\mathcal{L}_{\mathrm{IB}}$ \\
$\lambda_{1}$ & $10^{-3}$ & Weight of $\mathcal{L}_{\mathrm{reg}}$ \\
$\lambda_{2}$ & $10^{-5}$ & Weight of $\mathcal{L}_{\mathrm{align}}$ \\
$\lambda_{3}$ & $10^{-3}$ & Weight of $\mathcal{L}_{\mathrm{UCL}}$ \\
\hline
\end{tabular}
\caption{Hyperparameter settings used in the experiments.}
\label{tab:hyperparams_symbol}
\end{table}







\subsection{Main Results}

\begin{table}[!t]
\centering
\small
\setlength{\tabcolsep}{7pt}
\renewcommand{\arraystretch}{1.08}
\resizebox{\textwidth}{!}{
\begin{tabular}{llcccc}
\toprule
\textbf{Modality} & \textbf{Model} & \textbf{Acc(\%)} & \textbf{P(\%)} & \textbf{R(\%)} & \textbf{F1(\%)} \\
\midrule

\multirow{2}{*}{Image}
& ResNet~\cite{cai2019multi} & 64.76 & 54.51 & 70.80 & 61.53 \\
& ViT~\cite{dosovitskiy2020image}    & 67.83 & 57.93 & 70.07 & 63.43 \\
\midrule

\multirow{5}{*}{Text}
& Bi-LSTM~\cite{graves2005framewise} & 81.90 & 76.66 & 78.42 & 77.53 \\
& SIARN~\cite{tay2018reasoning}   & 80.57 & 75.55 & 75.70 & 75.63 \\
& SMSD~\cite{xiong2019sarcasm}    & 80.90 & 76.46 & 75.18 & 75.82 \\
& BERT~\cite{devlin2019bert}    & 83.85 & 78.72 & 82.27 & 80.22 \\
& RoBERTa~\cite{qin2023mmsd2} & 93.97 & 90.39 & \underline{94.59} & 92.45 \\
\midrule

\multirow{26}{*}{Multimodal}
& HFM~\cite{cai2019multi}                     & 86.63 & 83.84 & 84.18 & 84.01 \\
& D\&R Net~\cite{xu2020reasoning}            & 84.02 & 77.97 & 83.42 & 80.60 \\
& Bridge~\cite{wang2020building}             & 88.51 & 82.95 & 89.39 & 86.05 \\
& InCrossMGs~\cite{liang2021multi}           & 86.10 & 81.38 & 84.36 & 82.84 \\
& CMGCN~\cite{liang2022multi}                & 87.55 & 83.63 & 84.69 & 84.16 \\
& HKE-model~\cite{liu2022towards}            & 87.36 & 81.84 & 86.48 & 84.09 \\
& Att-BERT~\cite{pan2020modeling}            & 86.05 & 78.63 & 83.31 & 80.90 \\
& DIP~\cite{wen2023dip}                      & 89.59 & 87.76 & 86.58 & 87.17 \\
& KnowleNet~\cite{yue2023knowlenet}          & 88.87 & 88.59 & 84.18 & 86.33 \\
& DMSD-CL~\cite{jia2024debiasing}            & 88.95 & 84.89 & 87.90 & 86.37 \\
& AMIF~\cite{li2025ambiguity}                & 90.10 & 86.55 & 89.68 & 88.09 \\
& G2SAM~\cite{wei2024g}                      & 90.48 & 87.95 & 89.02 & 88.48 \\
& FSICN~\cite{lu2024fact}                    & 90.55 & 89.93 & 89.51 & 89.72 \\
& Multi-view CLIP~\cite{qin2023mmsd2}        & 88.33 & 82.66 & 88.65 & 85.55 \\
& MuMu~\cite{wang2024cross}                  & 90.73 & 88.81 & 88.44 & 88.62 \\
& MIL-Net~\cite{qiao2023mutual}              & 89.50 & 85.16 & 89.16 & 87.11 \\
& MICL~\cite{guo2025multi}                   & 92.08 & 90.05 & 90.61 & 90.33 \\
& DynRT-Net~\cite{tian2023dynamic}           & 93.59 & 93.06 & 93.60 & 93.31 \\
& LLaVA1.5~\cite{tang2024leveraging}         & 93.67 & \underline{93.70} & 93.14 & 93.40 \\
& LLaVA1.5-VIDR~\cite{tang2024leveraging}    & 89.97 & 89.26 & 89.58 & 89.42 \\
& DCPNet~\cite{fang2026dcpnet}               & 89.48 & 87.46 & 88.73 & 88.10 \\
& InterARM~\cite{yue2026interarm}            & 92.28 & 91.79 & 92.23 & 92.01 \\
& KFGC-Net~\cite{zhuang2025multi}            & 90.97 & 88.28 & 89.56 & 88.91 \\
& CIRM~\cite{zhao2025mmsd3}                  & 94.02 & 93.46 & 94.14 & 93.76 \\
& SCI-GDFN~\cite{xi2025multimodal}           & \underline{94.06} & 93.50 & 94.17 & \underline{93.80} \\
& \textbf{URMF (Ours)}                       & \textbf{95.02} & \textbf{94.69} & \textbf{95.19} & \textbf{94.91} \\
\bottomrule
\end{tabular}
}
\caption{Main results on the MSD dataset. The best results are shown in bold, and the second-best results are underlined.}
\label{tab:main_results_msd}
\end{table}

\begin{table}[!t]
\centering
\small
\setlength{\tabcolsep}{6pt}
\renewcommand{\arraystretch}{1.08}
\begin{tabular}{lcccc}
\toprule
\textbf{Model} & \textbf{Acc(\%)} & \textbf{P(\%)} & \textbf{R(\%)} & \textbf{F1(\%)} \\
\midrule
DynRT-Net~\cite{tian2023dynamic} & 71.40 & 71.80 & 72.17 & 71.34 \\
Multi-view CLIP~\cite{qin2023mmsd2} & \underline{85.64} & 80.33 & \textbf{88.24} & 84.10 \\
HKE-model~\cite{liu2022towards} & 76.50 & 73.48 & 72.07 & 72.25 \\
LLaVA1.5~\cite{tang2024leveraging} & 85.18 & \underline{85.89} & 85.20 & \underline{85.11} \\
Att-BERT~\cite{pan2020modeling} & 80.03 & 76.28 & 77.82 & 77.04 \\
\textbf{URMF} & \textbf{87.84} & \textbf{87.52} & 87.84 & \textbf{87.65} \\
\bottomrule
\end{tabular}
\caption{Main results on the MMSD2 dataset.}
\label{tab:main_results_mmsd2}
\end{table}

Table~\ref{tab:main_results_msd} and Table~\ref{tab:main_results_mmsd2} report the main results on MSD and MMSD2.
URMF achieves the best overall performance on both datasets, with \textbf{95.02\%} Accuracy on MSD and \textbf{87.84\%} Accuracy on MMSD2.
It outperforms representative unimodal, multimodal, and MLLM-based baselines, including strong recent methods such as CIRM and SCI-GDFN.
These results demonstrate the effectiveness and cross-dataset stability of uncertainty-aware robust fusion.

\subsection{Ablation Study}

\subsubsection{Component Ablation under Standard Settings}

To evaluate the contribution of each key component in URMF, we conduct ablation studies on the MSD and MMSD2 datasets.
Specifically, based on the full model, we separately remove the four training objectives ($\mathcal{L}_{\mathrm{align}}$, $\mathcal{L}_{\mathrm{IB}}$, $\mathcal{L}_{\mathrm{reg}}$, and $\mathcal{L}_{\mathrm{UCL}}$), the uncertainty-guided dynamic fusion mechanism, and the proposed cross-modal interaction order.
The results are shown in Table~\ref{tab:ablation_mmsd2}.
\begin{table}[!t]
\centering
\small
\setlength{\tabcolsep}{6pt}
\renewcommand{\arraystretch}{1.1}
\begin{tabular}{lcccc}
\toprule
\textbf{Variant} & \textbf{Acc(\%)} & \textbf{P(\%)} & \textbf{R(\%)} & \textbf{F1(\%)} \\
\midrule
w/o $\mathcal{L}_{\mathrm{align}}$ & 86.09 & 86.27 & 86.97 & 86.05 \\
w/o $\mathcal{L}_{\mathrm{IB}}$    & 86.67 & 86.80 & 87.51 & 86.62 \\
w/o $\mathcal{L}_{\mathrm{reg}}$   & 85.93 & 86.02 & 86.73 & 85.87 \\
w/o $\mathcal{L}_{\mathrm{UCL}}$   & 85.31 & 85.58 & 86.24 & 85.26 \\
w/o Dynamic Fusion                 & 86.43 & 86.26 & 86.94 & 86.33 \\
Standard Transformer               & 84.39 & 85.05 & 85.57 & 84.37 \\
\midrule
\textbf{URMF (Full)}               & \textbf{87.84} & \textbf{87.52} & \textbf{87.84} & \textbf{87.65} \\
\bottomrule
\end{tabular}
\caption{Ablation results of URMF on the MMSD2 dataset.}
\label{tab:ablation_mmsd2}
\end{table}
\begin{figure*}[!t]
    \centering
    \includegraphics[width=\textwidth]{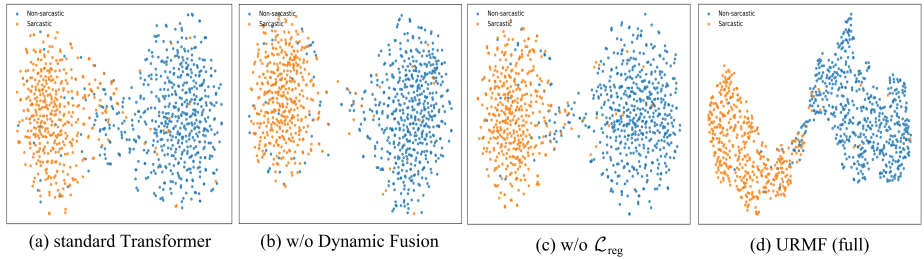}
    \caption{t-SNE visualization of the joint latent representations of major ablation variants and the full URMF on the MSD test set.}
    \label{fig:tsne_ablation}
\end{figure*}
As shown in Table~\ref{tab:ablation_mmsd2}, the full URMF achieves the best performance, reaching \textbf{87.84\%} Accuracy on MMSD2.
Removing any training objective, the dynamic fusion mechanism, or the proposed interaction order leads to performance degradation, indicating that all components contribute to the final model.
Among these variants, the Standard Transformer performs the worst, which shows the importance of the proposed ``cross-modal interaction first, intra-modal reasoning later'' design.
The visualization in Fig.~\ref{fig:tsne_ablation} further shows that the full URMF produces clearer class separation than the ablated variants, supporting the effectiveness of the proposed interaction and fusion design.

\subsubsection{Robustness Ablation under Unreliable Modality Conditions}

\begin{table}[t]
\centering
\small
\setlength{\tabcolsep}{5pt}
\renewcommand{\arraystretch}{1.12}
\begin{tabular}{l c c c cc}
\toprule
\multirow{2}{*}{\textbf{Model}} 
& \multicolumn{3}{c}{\textbf{Uncertainty Modeling}} 
& \multicolumn{2}{c}{\textbf{Unreliable Image}} \\
\cmidrule(lr){2-4} \cmidrule(lr){5-6}
& \textbf{Image} 
& \textbf{Text} 
& \textbf{Interaction} 
& \textbf{Acc(\%)} 
& \textbf{F1(\%)} \\
\midrule
ViT~\cite{dosovitskiy2020image} & -- & -- & -- & 58.78 & 58.78 \\
SCI-GDFN~\cite{xi2025multimodal} & -- & -- & -- & 85.26 & 85.24 \\
\midrule
\multirow{4}{*}{URMF}
&  & \cmark &  & 92.53 & 92.43 \\
& \cmark &  &  & -- & -- \\
& \cmark & \cmark &  & 93.69 & 93.58 \\
& \cmark & \cmark & \cmark & \textbf{94.77} & \textbf{94.67} \\
\bottomrule
\end{tabular}
\caption{Robustness and ablation results under the unreliable image setting.}
\label{tab:unreliable_image_ablation}
\end{table}
\begin{table}[t]
\centering
\small
\setlength{\tabcolsep}{5pt}
\renewcommand{\arraystretch}{1.12}
\begin{tabular}{l c c c cc}
\toprule
\multirow{2}{*}{\textbf{Model}} 
& \multicolumn{3}{c}{\textbf{Uncertainty Modeling}} 
& \multicolumn{2}{c}{\textbf{Unreliable Text}} \\
\cmidrule(lr){2-4} \cmidrule(lr){5-6}
& \textbf{Image} 
& \textbf{Text} 
& \textbf{Interaction} 
& \textbf{Acc(\%)} 
& \textbf{F1(\%)} \\
\midrule
RoBERTa~\cite{qin2023mmsd2} & -- & -- & -- & 75.09 & 74.99 \\
SCI-GDFN~\cite{xi2025multimodal} & -- & -- & -- & 76.71 & 76.59 \\
\midrule
\multirow{4}{*}{URMF}
&  & \cmark &  & -- & -- \\
& \cmark &  &  & 80.07 & 79.84 \\
& \cmark & \cmark &  & 81.11 & 80.91 \\
& \cmark & \cmark & \cmark & \textbf{82.65} & \textbf{82.36} \\
\bottomrule
\end{tabular}
\caption{Robustness and ablation results under the unreliable text setting.}
\label{tab:unreliable_text_ablation}
\end{table}
To evaluate robustness under unreliable modality conditions, we construct unreliable image and unreliable text settings by replacing a proportion of images or texts with noise on the MSD dataset.
As shown in Table~\ref{tab:unreliable_image_ablation} and Table~\ref{tab:unreliable_text_ablation}, the full URMF achieves the best performance in both settings, reaching \textbf{94.77\%} Accuracy under unreliable images and \textbf{82.65\%} Accuracy under unreliable text.
Compared with variants that model only partial modality uncertainty, the full model consistently performs better, indicating that uncertainty modeling over textual, visual, and interaction-aware representations is beneficial.
In particular, modeling the uncertainty of the interaction-aware representation further improves robustness, suggesting that it captures the stability of post-interaction incongruity cues rather than simply repeating unimodal uncertainty.

\section{Conclusion}

In this paper, we proposed URMF, an uncertainty-aware robust fusion framework for multimodal sarcasm detection.
URMF captures image-text incongruity through cross-modal interaction, models textual, visual, and interaction-aware representations with aleatoric uncertainty, and adaptively adjusts modality contributions according to estimated reliability.
Experiments on MSD and MMSD2 demonstrate that URMF achieves strong performance against representative unimodal, multimodal, and MLLM baselines.
Ablation and corrupted-modality robustness analyses further verify its effectiveness in preserving incongruity cues, suppressing unreliable evidence, and improving fusion robustness.
In future work, we will extend this framework to larger datasets and more complex multimodal understanding tasks.
\\
\\
\noindent\textbf{Acknowledgment.}
This work was supported by the Undergraduate Innovation and Entrepreneurship Training Program team led by Prof. Guoying Zhang at the School of Artificial Intelligence, China University of Mining and Technology-Beijing.

\bibliographystyle{splncs04}
\bibliography{references}

\end{document}